\documentclass{article}
\usepackage{ijcai24}
\usepackage{graphicx}
\usepackage{array}
\usepackage{times}
\usepackage{amsfonts}
\usepackage{soul}
\usepackage{url}
\usepackage[hidelinks]{hyperref}
\usepackage[utf8]{inputenc}
\usepackage[small]{caption}
\usepackage{graphicx}
\usepackage{amsmath}
\usepackage{amsthm}
\usepackage{booktabs}
\usepackage[table]{xcolor}
\usepackage{algorithm}
\usepackage{algorithmic}
\usepackage[switch]{lineno}
\usepackage{multirow}
\usepackage{booktabs,caption}
\usepackage[symbol]{footmisc}
\usepackage{tikz}
\usepackage{pifont}
\usepackage{diagbox}
\usepackage{makecell}
\newcommand*{\circled}[2][]{\tikz[baseline=(C.base)]{
    \node[inner sep=0pt] (C) {\vphantom{1g}#2};
    \node[draw, circle, inner sep=1pt, yshift=1pt] 
 at (C.center) {\vphantom{1g}};}}

\title{\texttt{\textbf{CLEAR}}: Cross-Transformers with Pre-trained Language Model is All you need for Person Attribute Recognition and Retrieval}

\usepackage{hyperref}
\newcommand{\printfnsymbol}[1]{%
\textsuperscript{\@fnsymbol{#1}}%
}
\author{
Doanh C. Bui$^1$\thanks{Equal contribution}
\and
Thinh V. Le$^2$$^*$\and
Ba Hung Ngo$^{3}$
\And
Tae Jong Choi$^3$\thanks{Corresponding author}\\
\affiliations
$^1$School of Electrical Engineering, Korea University, Republic of Korea\\
$^2$University of Information Technology, Ho Chi Minh Vietnam National University, Vietnam\\
$^3$Graduate School of Data Science, Chonnam National University, Republic of Korea\\
\emails
doanhbc@korea.ac.kr,
thinhlv.18@grad.uit.edu.vn,
ngohung@jnu.ac.kr,
ctj17@jnu.ac.kr
} 

\begin{document}
\maketitle

\begin{abstract}
Person attribute recognition and attribute-based retrieval are two core human-centric tasks. In the recognition task, the challenge is specifying attributes depending on a person's appearance, while the retrieval task involves searching for matching persons based on attribute queries. There is a significant relationship between recognition and retrieval tasks. In this study, we demonstrate that if there is a sufficiently robust network to solve person attribute recognition, it can be adapted to facilitate better performance for the retrieval task. Another issue that needs addressing in the retrieval task is the modality gap between attribute queries and persons' images. Therefore, in this paper, we present \texttt{\textbf{CLEAR}}, a unified network designed to address both tasks. We introduce a robust cross-transformers network to handle person attribute recognition. Additionally, leveraging a pre-trained language model, we construct pseudo-descriptions for attribute queries and introduce an effective training strategy to train only a few additional parameters for adapters, facilitating the handling of the retrieval task. Finally, the unified \texttt{\textbf{CLEAR}} model is evaluated on five benchmarks: PETA, PA100K, Market-1501, RAPv2, and UPAR-2024. Without bells and whistles, \texttt{\textbf{CLEAR}} achieves state-of-the-art performance or competitive results for both tasks, significantly outperforming other competitors in terms of person retrieval performance on the widely-used Market-1501 dataset.
\end{abstract} 

\section{Introduction}

Two of the most important human-centric tasks, which are person attribute recognition (PAR) and attribute-based person retrieval (AR), are crucial for real-world applications such as security or surveillance. While PAR can be considered a multi-label classification problem, AR involves the matching task between a person and a query attribute representation. Although these two problems are related to each other, previous studies \cite{tang2019improving,tan2020relation,chen2021enhance} tended to tackle them separately, overlooking the opportunity to unify these problems in one processing pipeline or model. Moreover, for the PAR problem, previous studies also commonly stuck with one standard CNN-based backbone, i.e., ResNet-50, and developed ad hoc modules exploring the characteristics of person images, which also missed other strong backbones that can effectively address the PAR problem without any specific human-centric modules \cite{cormier2023upar}. For the AR problem, various methods of image-textual matching \cite{li2017identity,dong2019person} or image-attribute matching techniques \cite{yin2017adversarial,cao2020symbiotic,jeong2021asmr} have been developed. However, the strength of embedding vectors for query attributes has not been fully explored.

\begin{figure}[!t]
\centerline{\includegraphics[width=0.49\textwidth]{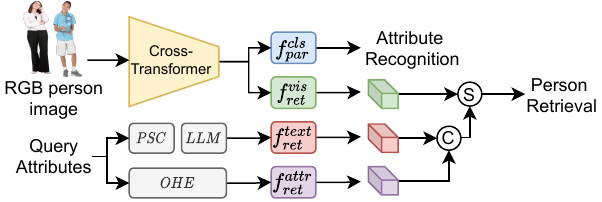}}
\caption{Unified \texttt{\textbf{CLEAR}} network for both person attribute recognition \& retrieval tasks. $f^{cls}_{par}$ denotes the head classifier for attribute recognition. $f^{vis}_{ret}$ denotes the auxiliary visual encoder. $f^{text}_{ret}$ denotes the auxiliary text encoder for the soft pseudo-description constructed from query attributes. $f^{attr}_{ret}$ denotes the auxiliary encoder for binary query attributes. \circled{C} represents the concatenation operation. \circled{S} represents the scoring for matching query attributes and persons during the search process.} \vspace{-0.8em}
\label{fig:general}
\end{figure}

In this study, we attempt to develop the strong unified model, called \texttt{\textbf{CLEAR}}: \textbf{C}ross-Transformers with Pre-trained \textbf{L}anguage is All you n\textbf{EE}d for Person \textbf{A}ttribute \textbf{R}ecognition and \textbf{R}etrieval. In summary, we have developed a robust two-branch cross-transformers backbone inspired by the vanilla vision transformer (ViT) \cite{dosovitskiy2020image} and swin transformer (SwinT) \cite{liu2021swin}. Vanilla vision transformer captures global long-range dependencies by computing self-attention on all patch tokens in a sequence, drawing inspiration from the vision transformer. Besides, our backbone incorporates elements from the swin transformer, enabling the learning of local long-range dependencies through the local computation of self-attention within windows. These windows are then shifted to ensure awareness of tokens across different windows. For robust learning of local long-range dependencies, we also introduce channel-aware self-attention before computing (shifted) window self-attention to boost important dimensions based on spatial information. At the head of our two-branch cross-transformers, a cross-fusion mechanism is presented to effectively aggregate two transformer-style branches, facilitating better attribute recognition performance.

The robust cross-transformers guarantee high performance in solving the PAR problem. Subsequently, we present a simple yet effective strategy for extending this approach to address the AR problem. Starting with query attributes, we introduce the concept of a pseudo description, which transforms discrete independent query attributes into a natural descriptive sentence. We then leverage a pre-trained GPT-based LLM to extract a strong representation, referred to as a \textit{soft embedding query}. Additionally, we utilize query attributes in binary form (represented by one-hot encoding), referred to as a \textit{hard embedding query}. Following this, we introduce lightweight learnable adapters and conduct margin learning for the image-attribute matching task. Figure \ref{fig:general} briefly represents our unified \texttt{\textbf{CLEAR}} model for person attribute recognition and attribute-based person retrieval.

In summary, our main contributions can be listed as below:
\begin{itemize}
\item We introduce a robust cross-transformers network that exploits both local-level and global-level long-range dependencies for the person attribute recognition task. The channel-aware self-attention is crafted to accentuate local-level features in high-level dimensions, and a cross-fusion module, based on attention mechanisms, is employed to aggregate two distinct types of long-range dependencies.

\item We propose the use of pseudo descriptions for attribute queries, represented by embeddings extracted from a powerful GPT-based pre-trained language model \cite{radford2018improving}.

\item We present an effective training strategy to extend the network designed for person attribute recognition to adapt to the attribute-based retrieval task. This approach results in the formation of a unified model for inference, referred to as \texttt{\textbf{CLEAR}}.

\item We evaluate the unified \texttt{\textbf{CLEAR}} model on published benchmarks: PA100K, PETA, RAPv2, Market-1501, and UPAR datasets. The results obtained from these evaluations, which achieve new state-of-the-art results for both human-centric tasks such as person attribute recognition and attribute-based retrieval tasks, serve as evidence of its effectiveness.

\end{itemize}
\section{Related Work}

\noindent\textbf{Person attribute recognition.} Previous approaches to person attribute recognition can be categorized into imbalance-aware techniques, attention-based mechanisms, multi-scale feature aggregation, and strong baselines. For imbalance-aware techniques, \cite{li2015multi} employed a weighted binary cross-entropy loss function and random image duplication to address imbalanced data distribution. In attention-based mechanisms, \cite{sarafianos2018deep} and \cite{liu2017hydraplus} utilized visual attention, with \cite{sarafianos2018deep} aggregating attention masks and \cite{liu2017hydraplus} applying multi-level fusion with visual semantic attributes. \cite{li2018pose} explored pose information using the Spatial Transformer Network (STN) \cite{jaderberg2016spatial}. In multi-scale features aggregation, \cite{tan2020relation} proposed an end-to-end model with GCN-based modules, while \cite{tang2019improving} presented a framework with ALM and FPN modules. \cite{zhong2021improving} addressed distance-related drops with MSSC, incorporating non-local attention and long-range dependencies. Considering person attribute recognition as a multi-label classification problem, \cite{specker2023upar} introduced a simple and strong framework with a ConvNeXt backbone and enhancements like exponential moving averages, suitable batch sizes, label smoothing, dropout, and data augmentation.

\noindent\textbf{Person attribute-based retrieval.} The goal of person attribute-based retrieval is to align image embeddings and attribute descriptions in a joint cross-modal feature space, with numerous proposed approaches employing diverse techniques. These approaches can be classified into adversarial learning, zero-shot learning, and attention learning. \cite{yin2017adversarial} and \cite{cao2020symbiotic} leveraged adversarial learning to enhance semantic consistency across modalities. While \cite{yin2017adversarial} used adversarial learning that enables query attributes and image features to match at both global and semantic levels, \cite{cao2020symbiotic} utilized generative adversarial networks
(GANs) that collaborates to mutually benefit each other for optimizing the cross-modal alignment on the common embedding space. For zero-shot learning, \cite{dong2019person} and \cite{jia2021rethinking} introduced new perspectives of the person attribute retrieval task under the zero-shot learning setting. \cite{dong2019person} presented a deep learning model named Attribute-Image Hierarchical Matching (AIHM), which is able to increase the reliability of matching text attribute descriptions under noisy surveillance with visual embeddings. In attention learning, \cite{li2017identity} introduced a new training strategy with multiple stages and a latent co-attention mechanism to efficiently alleviate incorrect textual and visual matching. 

\section{Methodology}

\subsection{Overview}

In this section, we present in detail how the unified \texttt{\textbf{CLEAR}} network is designed for two tasks: 1) \textbf{person attribute recognition} and 2) \textbf{attribute-based person retrieval}. For the recognition task, given an image $I \in \mathbb{R}^{H\times W \times 3}$, the \texttt{\textbf{CLEAR}} network will produce the prediction $\mathbf{\hat{y}} = \{\hat{y}_i\}^{N_{attr}}_{i=1}$, where $\mathbf{\hat{y}} \in \mathbb{R}^{\{0, 1\}}$, and $i$ denotes the predicted $i^{\text{th}}$ attribute. For the retrieval task, \texttt{\textbf{CLEAR}} receives the query $\mathbf{q} = \{q_i\}^{N_{attr}}_{i=1}$, where $\mathbf{q} \in \mathbb{R}^{\{0, 1\}}$, $i$ denotes the $i^{\text{th}}$ attribute that needs to perform retrieval, and outputs a set of $k$ person images $\mathbf{\hat{I}} = \{I_i\}^{k}_{i=1}$ that have the same attributes as the query. \texttt{\textbf{CLEAR}} consists of sub-networks: $f_{par}(\cdot,\theta_{par})$ and $f_{ret}(\cdot,\theta_{ret})$. $f_{par}(\cdot,\theta_{par})$ is the strong cross-transformers network used to solve the person attribute recognition task. $f_{ret}(\cdot,\theta_{ret})$ is a very lightweight network attached to $f_{par}(\cdot,\theta_{par})$ for training the attribute-based retrieval task. In this case, $f_{par}(\cdot,\theta_{par})$ serves as the visual encoder, where $\theta_{par}$ remains non-updated during training the retrieval task.

\subsection{Cross-Transformers for attribute recognition} \label{sec:cross-trans}

\begin{figure}[!t]
\centerline{\includegraphics[width=0.35\textwidth]{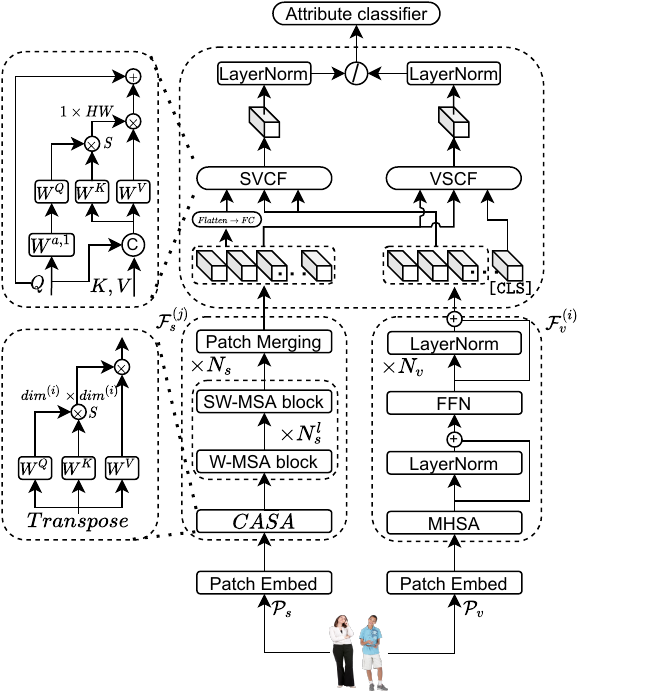}}
\caption{Cross-Transformers backbone ($f_{par}$) for person attribute recognition.} \vspace{-0.8em}
\label{fig:framework}
\end{figure}

\textbf{Transformer-style networks.} Both SwinT and Vanilla ViT share learning behaviors, processing the input image as a sequence of patch tokens. ViT's core includes a multi-head self-attention (MSA) mechanism and feedforward neural networks (FFN), while SwinT uses multi-layer shifted windows (SW) self-attention to benefit multiple-layer perception (MLP). The key difference lies in self-attention computation; ViT processes non-overlapping fixed-size patches, while SwinT employs hierarchical representation with gradually merging neighboring patches. Herein, we observe that Vanilla ViT captures global-level long-range dependencies involving all tokens in self-attention. In contrast, SwinT focuses on local-level long-range dependencies, computing self-attention within local (shifted) windows. To leverage both dependencies, we design a strong cross-transformer backbone. In Figure \ref{fig:framework}, for a person image $I \in \mathbb{R}^{3 \times H \times W}$, the two-branch cross-transformers (we will denote as $f_{par}$ in the context of retrieval task in below sections) process can be formulated as follows:
\begin{equation}
\renewcommand{\arraystretch}{1.3}
\resizebox{0.4\textwidth}{!}{
$\begin{array}{ll}
z^{(0)}_v = \mathcal{P}_v(I), z^{(0)}_s = \mathcal{P}_s(I), \\
z^{(i)}_v = \mathcal{F}_{v}^{(i)}(z^{(i-1)}_v), 1 \leq i \leq N_v \\
z^{(j)}_s = \mathcal{F}_{s}^{(j)}(CASA(z^{(j-1)}_s)), 1 \leq j \leq N_s \\
\mathbf{z}_s = LN_s(SVCF(FC(Flatten(z^{(N_s)}_s)), z^{(N_v)}_v)), \\
\mathbf{z}_v = LN_v(VSCF(z^{(N_v), \texttt{[CLS]}}_v, z^{(N_s)}_s)), \\
\mathbf{z} = Mean(\mathbf{z}_s, \mathbf{z}_v),
\end{array}$
}
\end{equation}

\noindent Let $\mathcal{P}_v$ and $\mathcal{P}_s$ be patch embedding networks that process input image $I$ into sequences of tokens $z^{(0)}_v \in \mathbb{R}^{p^2_v \times dim^{v}_{emb}}$ and $z^{(0)}_s \in \mathbb{R}^{p^2_s \times dim^{s,0}_{emb}}$. $\mathcal{F}^{(j)}_s$ represents the $j^{th}$ Swin Transformer block with $N^l_s$ layers, each containing (S)W-MSA and an MLP network, preceded by layer norm. A skip connection merges the input sequence with the (S)W-MSA-generated sequence before MLP. W-MSA and SW-MSA are interleaved. $\mathcal{F}^{(i)}_v$ is the $i^{th}$ Transformer block with layer normalization, MHSA, and a feedforward network. $Flatten$ flattens output tokens from final SwinT layer into a vector, followed by $FC$ projection to a lower dimension, $dim^{v}_{emb}$ in $\mathcal{F}_{v}$. $CASA$ is channel-aware self-attention, while $SVCF$ and $VSCF$ are cross self-attentions between Transformer branches, detailed in subsequent sections. $LN_s$ and $LN_v$ are independent normalization layers capturing different distributions from the two Transformer branches.

\noindent\textbf{Channel-aware self-attention.} The core of transformer-style models lies in the (multi-head) self-attention mechanism \cite{vaswani2017attention}, forming connections among all patch tokens in the sequence $\mathbf{z} = \{z_i\}^{N}_{i=1}$. Matrices $\mathbf{W}^q$, $\mathbf{W}^k$, and $\mathbf{W}^v$ are applied to $\mathbf{z}$ to produce query ($\mathbf{q}$), key ($\mathbf{k}$), and value ($\mathbf{v}$) tokens. The attention weight matrix, computed by the scaled dot product between $\mathbf{q}$ and $\mathbf{v}$ followed by softmax, is denoted as $att$. This is then multiplied with $\mathbf{v}$ to highlight specific tokens.

Despite its spatial attention capabilities, self-attention does not inherently consider channel relationships, often leading to their oversight. To overcome this limitation, we introduce channel-aware self-attention ($CASA$) at the top of each SwinT block. Given the output $z^{i}_s \in \mathbb{R}^{dim_{emb}^{s,i} \times h^{(i)} \times w^{(i)}}$ at the $i^{th}$ SwinT block, $dim_{emb}^{s,i}$ represents  the embedding size at $i^{th}$ SwinT block, $CASA$ is defined as an operation, denoted as ($CASA(\cdot)$), involving the following steps:
\begin{equation}
\renewcommand{\arraystretch}{1.3}
\resizebox{0.4\textwidth}{!}{$\begin{array}{ll}
z^{i,T}_s = \text{Transpose}(\text{Group}(z^{i}_s)), z^{i,T}_s \in \mathbb{R}^{(h^{(i)} w^{(i)}) \times dim_{emb}^{s,i}}, \\
\mathbf{q}^{i} = z^{i,T}_s \odot \mathbf{W}^{q,i}, \mathbf{k}^{i} = z^{i,T}_s \odot \mathbf{W}^{k,i}, \mathbf{v}^{i} = z^{i,T}_s \odot \mathbf{W}^{v,i}, \\
\text{where } \mathbf{W}^{\cdot,i} \in \mathbb{R}^{h^{(i)}w^{(i)} \times h^{(i)} w^{(i)}}, \\
att = \text{softmax}\big(\frac{\mathbf{q}^{i}{\mathbf{k}^{i}}^T}{\sqrt{d_k}}\big), att \in \mathbb{R}^{h^{(i)}w^{(i)} \times h^{(i)} w^{(i)}}, \\
z^{i,sb,T}_s = z^{i,T}_s + att \odot z^{i,T}_s, \\
z^{i,sb}_s = \text{Ungroup}(\text{Transpose}(z^{i,sb,T}_s)), \\ z^{i,sb}_s \in \mathbb{R}^{dim \times h^{(i)} \times w^{(i)}}, \\
z^{i+1}_s = \mathcal{F}_{s}^{(i)}(z^{i,sb}_s),
\end{array}$}
\end{equation}
\noindent where \(z^{i,T}_s\) denotes the transpose matrix of Group(\(z^{i}_s\)), with the group of spatial resolutions \((h^{(i)} \times w^{(i)})\) considered as rows of the matrix. Subsequently, we compute the channel-aware attention weight matrix \(att\) to identify crucial features, based on the spatial tokens. Finally, \(att\) is multiplied with the transpose matrix \(z^{i,T}_s\) along with the skip connection, resulting in \(z^{i,sb,T}_s\). This is then rearranged to \(z^{i,sb}_s\), which represents the input to the \(i^{th}\) SwinT block $\mathcal{F}^{(i)}_s$.

\noindent\textbf{Cross-fused self-attention.} At the top of the two network branches, we obtain feature vectors $\mathbf{z}^s \in \mathbb{R}^{dim_{emb}^{s,N_s}}$ and $\mathbf{z}^v \in \mathbb{R}^{dim_{emb}^{v}}$, along with sequences of tokens $z^{(N_s)}_s \in \mathbb{R}^{dim_{emb}^{s,N_s} \times h^{(N_s)} \times w^{(N_s)}}$ and $z^{(N_v)}_v \in \mathbb{R}^{N^{tok}_v \times dim_{emb}^{v}}$, $N^{tok}_v$ is the number of tokens in Vanilla ViT. These represent the SwinT feature vector compressed by $Flatten$ followed by an $FC$ layer, the \texttt{[CLS]} token from Vanilla ViT, and sequences of tokens from SwinT and Vanilla ViT, respectively.

Although concatenating \(\mathbf{z}^s\) and \(\mathbf{z}^v\) demonstrates effectiveness, it underutilizes \(z^{(N_v)}_v\) and \(z^{(N_s)}_s\), which contain valuable features. Given \(\mathbf{z}^s\) and \(z^{(N_v)}_v\), we design cross-attention, inspired by \cite{chen2021crossvit}, as follows:

\begin{equation}
\renewcommand{\arraystretch}{1.3}
\resizebox{0.4\textwidth}{!}{$\begin{array}{ll}
\hspace{-4.3mm}
\mathbf{z}^{s,a} = \mathbf{z}^{s}\odot\mathbf{W}^{a,1}, \mathbf{W}^{a,1} \in \mathbb{R}^{dim \times dim} \\
\hspace{-4.3mm}
\mathbf{z}^{s,c} = \text{Concatenate}(\mathbf{z}^{s,a}, z^{(N_v)}_v), \mathbf{z}^{s,c} \in \mathbb{R}^{(h_vw_v+1) \times dim}\\
\hspace{-4.3mm}
\mathbf{q} = \mathbf{z}^{s,a}\odot\mathbf{W}^{q}, \mathbf{k} = \mathbf{z}^{s,c}\odot\mathbf{W}^{k}, \mathbf{v} = \mathbf{z}^{s,c}\odot\mathbf{W}^{v}, \\
\hspace{-4.3mm}
att = \text{softmax}(\frac{\mathbf{q}\mathbf{k}^T}{\sqrt{d_k}}), att \in \mathbb{R}^{1\times(h_vw_v+1)}, \\
\hspace{-4.3mm}
\mathbf{z}^{s'} = (\mathbf{z}^{s,c} + \mathbf{z}^{s,c} \odot att) \odot \mathbf{W}^{a,2}, \mathbf{W}^{a,2} \in \mathbb{R}^{dim \times dim},
\end{array}$}
\label{eq:ca}
\end{equation}

\noindent where $\mathbf{W}^{a,1}$ and $\mathbf{W}^{a,2}$ denotes  the projection matrices for dimension alignment. $\mathbf{z}^{s,a}$ denotes aligned token features, and $\mathbf{z}^{s,c}$ represents the concatenation of $\mathbf{z}^{s,a}$ and $z^{(N_v)}_v$. This approach allows a single token from SwinT to compute self-attention with a sequence of tokens from Vanilla ViT, producing attention weights $att \in \mathbb{R}^{1 \times (h_vw_v+1)}$. This method, called the \(SVCF(\cdot)\) operation, makes a token aware of other sequences, compressing knowledge across channels and explaining how $\mathbf{z}^s$ becomes aware of $z^{(N_v)}_v$. The \(VSCF(\cdot)\) operation is applied for $\mathbf{z}^v$ to be aware of $z^{(N_s)}_s$. However, $z^{(N_s)}_s \in \mathbb{R}^{dim \times h^{(N_s)} \times w^{(N_s)}}$ is initially in tensor form, requiring grouping of spatial resolutions as a token sequence for cross-fusion with $\mathbf{z}^v$ through a cross-attention mechanism.

\noindent\textbf{Independent Layer Normalization.} Utilizing two transformer-style networks with distinct perspectives, we introduce independent layer normalizations before fusion, tailored for representations $\mathbf{z}^{s'}$ and $\mathbf{z}^{v'}$. This is expressed as:
\begin{equation}
\renewcommand{\arraystretch}{1.3}
\begin{array}{ll}
\mathbf{M}^{s} = LN_{s}(\mathbf{z}^{s'}; \alpha_s, \beta_s),
\mathbf{M}^{v} = LN_{v}(\mathbf{z}^{v'}; \alpha_v, \beta_v),
\end{array}
\label{eq:ca}
\end{equation}

\noindent where $LN_{s}$ and $LN_{v}$ are layer normalizations for output from $\mathcal{F}_{s}$ i.e., $\mathbf{z}^{s'}$ and output from $\mathcal{F}_{v}$, i.e., $\mathbf{z}^{v'}$. Learnable scale and shift parameters, $\alpha_s$, $\beta_s$, $\alpha_v$, and $\beta_v$, are applied to the affine feature values.

We then concatenate the normalized representations, pass them through a fully-connected layer, and obtain $N_{attr}$-dimensional logits. Finally, applying the $\sigma(\cdot)$ function yields the final output for multi-label classification:
\begin{equation}
\renewcommand{\arraystretch}{1.3}
\begin{array}{ll}
\mathbf{M}^{sv} = FC\big(Concat(\mathbf{M}^{s}, \mathbf{M}^{v})\big), \hat{\mathbf{y}} = \sigma(\mathbf{M}^{sv}).
\end{array}
\label{eq:ca}
\end{equation}

\noindent\textbf{Loss function.} To facilitate multi-label classification, we employ the binary cross-entropy (BCE) loss function for supervised learning. Considering \(\hat{\mathbf{y}} \in \mathbb{R}^{N_{attr}}\) as the output from the \(\sigma(\cdot)\) function and \(\mathbf{y} = \{y_i\}_{i=1}^{N_{attr}} \in \mathbb{R}^{N_{attr}}\) as the one-hot encoding ground-truth, where \(y_i \in \{0, 1\}\), the binary cross-entropy loss function is defined as follows:
\begin{equation}
\mathcal{L}_{BCE}(\hat{\mathbf{y}}, \mathbf{y}) = -\left[\mathbf{y}\log(\hat{\mathbf{y}}) + (1 - \mathbf{y})\log(1 - \hat{\mathbf{y}})\right].
\end{equation}

\subsection{Language-based margin learning for retrieval} \label{sec:soft-pseudo}
\textbf{Soft-hard embedding queries.} Given query attributes $\mathbf{q} = \big\{q_i\big\}^{N_{attr}}_{i=1}$, where $q_i$ is $i^{th}$ attribute in word form, we construct the soft pseudo description having fixed \(N_w\) as follows:

\begin{figure}[!t]
\centerline{\includegraphics[width=0.45\textwidth]{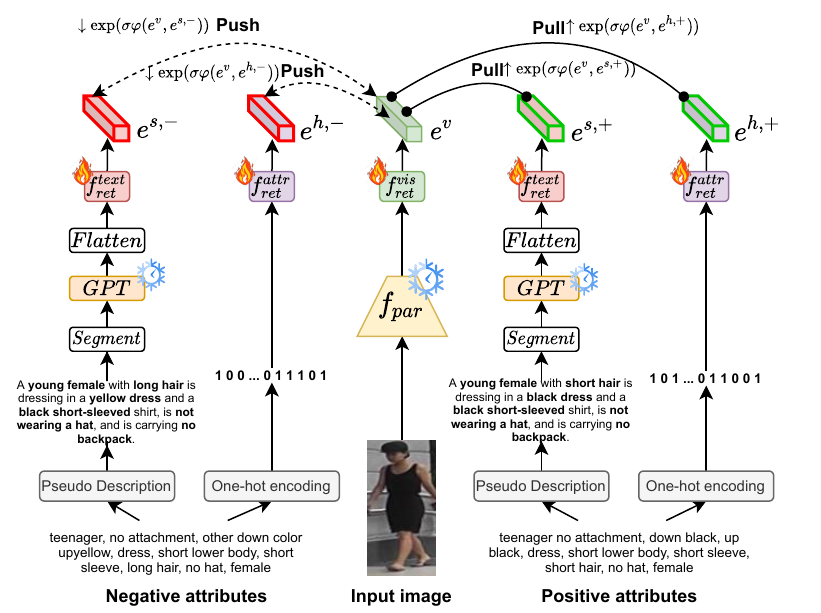}}
\caption{Margin learning with pseudo description (\textit{soft embedding query}) and hard binary attribute (\textit{hard embedding query}).} \vspace{-1.5em}
\label{fig:framework_pseudo}
\end{figure}

\begin{equation}
\resizebox{0.35\textwidth}{!}{$\begin{array}{ll}
\text{This is a photo of } \colorbox{blue!30}{\texttt{\textbf{[AGE]}}} \text{ } \colorbox{green!30}{\texttt{\textbf{[GENDER]}}} \text{ \underline{taken from} } \\ \colorbox{yellow!30}{\underline{\texttt{\textbf{[CAMERA ANGLE]}}}} \text{ with } \colorbox{purple!30}{\texttt{\textbf{[HAIR LENGTH]}}} \text{ hair, is dressed } \\ \text{in a }  \colorbox{orange!30}{\texttt{\textbf{[LOWER COLOR]}}} \colorbox{gray!30}{\texttt{\textbf{[LOWER BODY CLOTHING]}}} \text{ and a } \\ \colorbox{pink!30}{\texttt{\textbf{[UPPER COLOR] }}} \colorbox{red!30}{\texttt{\textbf{[UPPER BODY CLOTHING]}}} \\ \text{ with } \colorbox{brown!30}{\texttt{\textbf{[SLEEVES LENGTH]}}} \text{ sleeves } \text{ \underline{with} } \\ \colorbox{cyan!30}{\underline{\texttt{\textbf{[UPPER BODY MOTIF]}}}} \text{ \underline{motif}} \text{, is } \colorbox{lime!30}{\texttt{\textbf{[WEARING]}}} \\ \text{ a } \colorbox{teal!30}{\texttt{\textbf{[ACCESSORY]}}} \text{, carrying } \colorbox{olive!30}{\texttt{\textbf{[BAG]}}} \text{ \underline{and} } \colorbox{violet!30}{\underline{\textbf{\texttt{[BACKPACK]}}}}.
\end{array}$}
\label{eq:pseudo}
\end{equation}

\noindent where a set of \texttt{[TAG]} is filled out based on set of query attributes $\mathbf{q}$. Some tags/auxiliary terms are underlined, indicating differences in query attributes between datasets. In these cases, we simply add or remove these tags/terms to adapt the datasets for training/evaluation. Figure \ref{fig:framework_pseudo} illustrates how the pseudo-description is constructed from attributes. Herein, for discrete binary attributes, auxiliary terms (i.e., with,``\textit{hair}," ``\textit{dressing}," ``\textit{carrying}") are utilized. We believe that these terms effectively connect attributes together, forming a meaningful description. Next, we tokenize the sentence into a set of words $\mathbf{w}=\{w_i\}^{N_w}_{i=1}$ and leverage a strong pre-trained language model to transfer it to the embedding space. This results in a sequence of embedding vectors $\{s_i\}^{N_w}_{i=1}$, where $s_i \in \mathbb{R}^{dim_w}$ and $dim_w$ is dimension size. To represent the sequence of embedding vectors as a unique embedding vector, we flatten all $e_i$ as one single $N_w \times dim_w$-dimensional vector: $\mathbf{s}=\text{Flatten}\big(\{s_i\}^{N_w}_{i=1}\big)$, where $\mathbf{s} \in \mathbb{R}^{(N_{w}\times dim_w)}$. Because $\mathbf{s}$ is built based on a pseudo caption constructed by using auxiliary terms which are described in (\ref{eq:pseudo}), we refer it as \textit{soft embedding query}.

Besides, while $\mathbf{s}$ is a strong, meaningful embedding vector inspired by natural language linking, discrete one-hot encoded binary attributes $\mathbf{h} = \{h_i\}^{N_{attr}}_{i=1}, h_i \in \mathbb{R}^{\{0,1\}}$ still provide useful information in the embedding space. Hence, we also leverage query attributes in binary form to enrich the final query embedding vector, facilitating improved search performance. We refer to binary attribute query as \textit{hard embedding query}.

\noindent\textbf{Margin learning strategy.} Given the trained Cross-transformers backbone $f_{par}$, to adapt to the retrieval task, we only introduce lightweight, learnable adapters $f^{vis}_{ret}$, $f^{text}_{ret}$, and $f^{attr}_{ret}$ to produce embeddings for person images, pseudo descriptions, and binary query attributes, respectively. All of adapters has the same architecture which is a stack of three linear projections, and each followed by ReLU activation, except for the last projection, to produce encoded embedding vector. $f^{vis}_{ret}$ produces the encoded vector has embedding size of $dim^{vis}_{emb}$, while $f^{attr}_{ret}$ and $f^{text}_{ret}$ produces the encoded vectors have embeding size of $dim^{query}_{emb}$, where $dim^{query}_{emb} = dim^{vis}_{emb}/2$. Note that $f_{par}$ is frozen during margin learning. Following \cite{deng2019arcface,jeong2021asmr}, the objective of margin learning is to pull embeddings of person images toward their corresponding attribute embeddings and push them away from other attribute embeddings. Given two set of embeddings $\mathbf{f}$ and $\mathbf{g}$ that need to be performed margin learning, we adopt the margin loss introduced in \cite{deng2019arcface}:

\begin{equation}
\begin{array}{ll}
\mathcal{L}_{MA} = -\log\Bigg(
\frac{\exp(\sigma\varphi(\mathbf{f},\mathbf{g}^{+}))}{\exp(\sigma\varphi(\mathbf{f},\mathbf{g}^{+}) ) + \sum_{\mathbf{g}^{-}}\exp(\sigma\varphi(\mathbf{f},\mathbf{g}^{-}))}\Bigg), \\
\varphi(\cdot,\cdot) = \sigma\cos\big(\alpha(\mathbf{f}, \mathbf{g}) + \gamma\big),
\end{array}
\end{equation}

\noindent where $\mathbf{g}^{+}$ represents attribute embeddings of corresponding person embeddings $\mathbf{f}$ that need to be pulled towards each other. $\mathbf{g}^{-}$ represents the set of other attribute embeddings that will be learned to push away from $\mathbf{f}$. $\varphi$ denotes the variant of the cosine similarity function between the sets of embeddings $\mathbf{f}$ and $\mathbf{g}$, which includes the scale factor $\sigma$ and the modality margin factor $\gamma$. $\alpha$ denotes the angle formed by two embeddings. Given the set of person images $\mathbf{I} = \{I\}^{N_i}_{i=1}$, the set of \textit{hard query embedding} vectors $\mathbf{H}=\{\mathbf{h}_i\}^{N_{q}}_{i=1}$, and the set of \textit{soft query embedding} vectors $\mathbf{S}=\{\mathbf{s}_i\}^{N_{q}}_{i=1}$, learnable adapter $f^{vis}_{ret}$ (following the frozen cross-transformers backbone $f{par}$) is used to extract a visual-encoded person embedding with an embedding size of $dim^{vis}_{emb}$. In addition, $f^{text}_{ret}$ and $f^{attr}_{ret}$ are utilized to extract sets of encoded embedding vectors for \textit{soft query embedding} and \textit{hard query embedding}, respectively, both with an embedding size of $dim^{query}_{emb}$

\begin{equation}
\begin{array}{ll}
\mathbf{E}^{v} = f^{vis}_{ret}\big(f_{par}(\mathbf{I})\big), \mathbf{E}^{h} = f^{attr}_{ret}(\mathbf{H}), \mathbf{E}^{s} = f^{text}_{ret}(\mathbf{S}), \\
\end{array}
\end{equation}

\noindent where $\mathbf{E}^{p}$, $\mathbf{E}^{h}$ and $\mathbf{E}^{s}$ are sets of encoded embeddings for sets of person images, hard query embeddings and soft query embeddings, respectively. Then, margin learning is performed for two pairs $<\mathbf{E}^{p}, \mathbf{E}^{h}>$ and $<\mathbf{E}^{p}, \mathbf{E}^{s}>$:

\begin{equation}
\mathcal{L}_{total} = \beta_1\mathcal{L}_{MA}(\mathbf{E}^{p}, \mathbf{E}^{h}) + \beta_2\mathcal{L}_{MA}(\mathbf{E}^{p}, \mathbf{E}^{s}),
\label{eq:loss_total}
\end{equation}

\noindent where $\beta_1$ and $\beta_2$ represent the weights for two loss terms.

After training, $\mathbf{E}^h \in \mathbb{R}^{N_q \times dim^{query}_{emb}}$ and $\mathbf{E}^s \in \mathbb{R}^{N_q \times dim^{query}_{emb}}$ are concatenated, and similarity scores are computed with $\mathbf{E}^v$. The retrieval results are then sorted by the order of their score values.

\section{Experimental Results}

\begin{table}[]
\centering
\resizebox{0.42\textwidth}{!}{\begin{tabular}{l|c|c|c|c|c|c}\toprule
\textbf{Datasets}     & \textbf{PETA} & \textbf{Market-1501} & \textbf{PA100K} & \textbf{RAPv2} & \multicolumn{2}{c}{\makecell{\textbf{UPAR2024} \\ \textbf{(dev/official test)}}} \\ \midrule
\# Attribute & 65  & 27      & 26 & 72 & \multicolumn{2}{c}{40}   \\
\# Group     & 17  & 10      & 15 & 16 & \multicolumn{2}{c}{12}   \\ \midrule
\# Train person category & 5858& 508     & 2020 & -- & \multicolumn{2}{c}{5237} \\
\# Train image  & 15067        & 12936   & 80000  & 67943        & \multicolumn{2}{c}{97669}\\ \midrule
\# Test person category  & 1552& 484     & 849& -- & 2738$^{*}$ & 351 \\
\# Unseen category       & 1242& 315     & 168& -- & 799$^{*}$  & 151\\
\# Test image& 3933    & 16483   & 10000 & 16985   & 33407$^{*}$ & 28095\\
\bottomrule
\end{tabular}}
\\
\footnote[1]{} \footnotesize Statistics for UPAR2024 dev-test.
\caption{Statistics of five benchmarks.} \vspace{-0.8em}
\label{tab:stas}
\end{table}

\begin{figure}[!t]
\centerline{\includegraphics[width=0.4\textwidth]{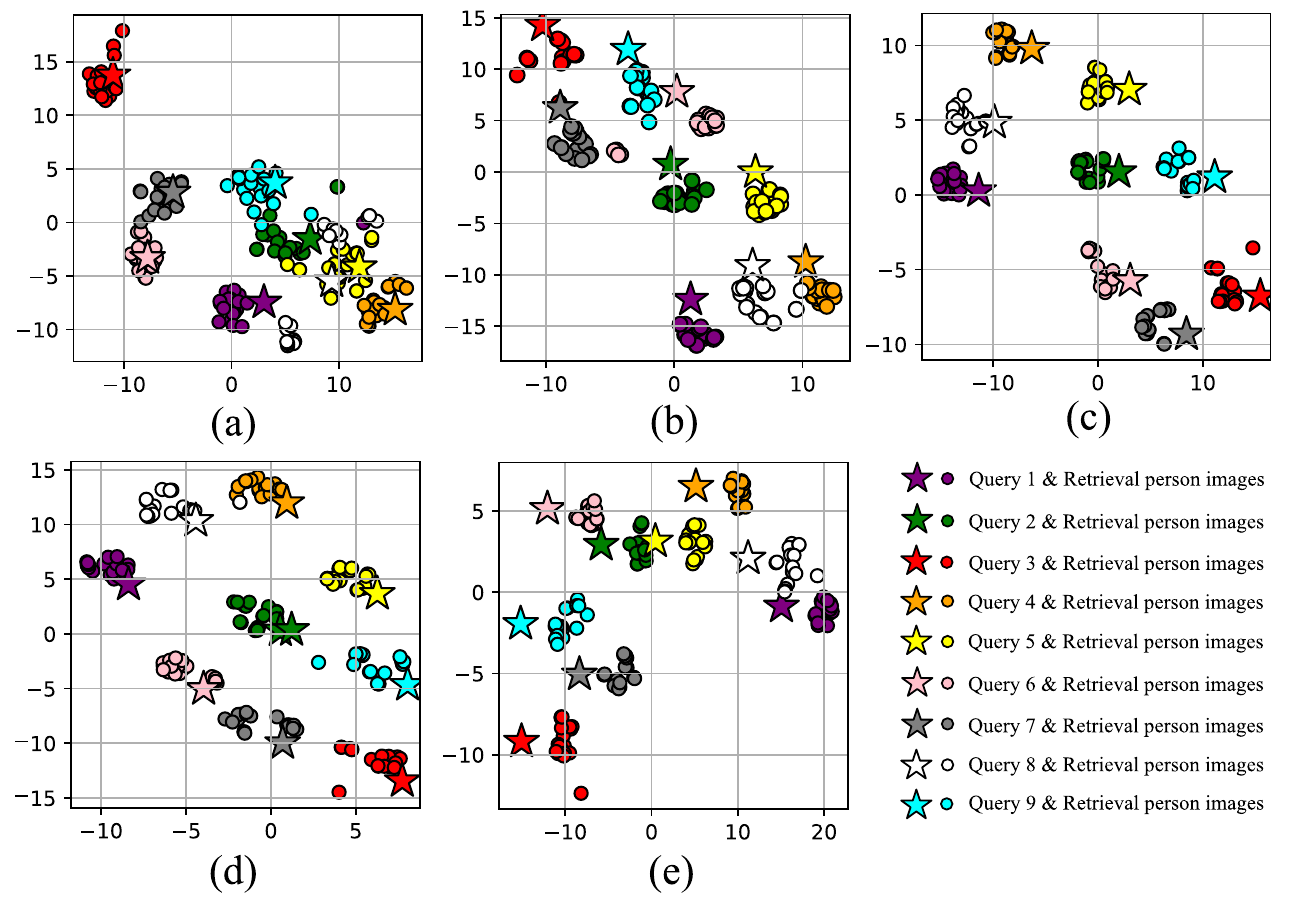}} \vspace{-0.8em}
\caption{t-SNE visualization for ten randomly chosen queries, with each query accompanied by its corresponding set of 20 person images. $\star$ denotes query representations. $\circ$ denotes person image representations. (a) ASMR (b) Hard Binary Attribute (\textbf{HA}). (c) Attribute Word (\textbf{W}). (d) Soft Pseudo Caption (\textbf{SP}). (e) Soft + Hard Query (\textbf{Ours}).} \vspace{-0.8em}
\label{fig:tsne}
\end{figure}

\begin{table*}[!h]
\centering
\resizebox{0.83\textwidth}{!}{\begin{tabular}{c|c|cc|cc|cc|cc|cc}
\toprule
\multirow{2}{*}{\textbf{Methods}} & \multirow{2}{*}{\textbf{Backbone}} & \multicolumn{2}{c}{\textbf{PA100K}} & \multicolumn{2}{c}{\textbf{PETA}} & \multicolumn{2}{c}{\textbf{RAPv2}} & \multicolumn{2}{c}{\textbf{Market-1501}} & \multicolumn{2}{c}{\makecell{\textbf{UPAR2024} \\ \textbf{dev-test}}} \\ \cmidrule{3-12}
 & & mA & F1& mA& F1 & mA& F1& mA & F1& mA& F1 \\ \midrule
ALM (ICCV'19) \cite{tang2019improving}& BN-Inception& 80.7 & 86.5& 86.3& 86.9 & 78.2& 77.3& 78.0 & 84.9& 82.6& 85.5 \\
VAC (CVPR'19) \cite{guo2019visual}& ResNet50& 79.0 & 86.8& 83.6& 86.2 & --& --& -- & --& --& -- \\
MSCC (IJCNN'21) \cite{zhong2021improving} & ResNet50& 82.1 & 86.8& 80.8& 87.4 & 80.2& 79.1& 78.8 & 83.0& \textbf{\textcolor{blue}{84.1}}$^{*}$ & 85.7$^{*}$ \\
VFA (ICCV'21) \cite{chen2021enhance}& ResNet50& 81.3 & 87.0& 86.5& 87.3 & --& --& -- & --& --& -- \\
JLAC (AAAI'20) \cite{tan2020relation} & ResNet50& 82.3 & 87.6& 87.0& 87.5 & --& --& -- & --& --& -- \\
Strong baseline \cite{jia2021rethinking} & ResNet50 & 84.0 & 86.3 & 81.6 & 88.1 & 77.4 & 78.5 & 76.5 & 83.6 & 82.3$^{*}$ & \textbf{\textcolor{blue}{86.4}}$^{*}$ \\
DAFL (AAAI'22) \cite{jia2022learning} & ResNet50 & 83.5 & 88.1 & 87.1 & 86.0 & \textbf{\textcolor{blue}{81.0}} & 79.1 & -- & -- & -- & -- \\
UPAR (WACV'23) \cite{specker2023upar} & ResNet50& 82.2 & 88.5& 87.1& 87.7 & 78.8& 80.0& 79.5 & 85.4& --& -- \\
UPAR (WACV'23) \cite{specker2023upar} & ConvNeXt-B& \textbf{\textcolor{blue}{84.8}} & \textbf{\textcolor{blue}{90.2}} & \textbf{\textcolor{red}{88.4}}& \textbf{\textcolor{red}{89.9}} & 79.9& \textbf{\textcolor{blue}{81.0}} & \textbf{\textcolor{blue}{81.5}} & \textbf{\textcolor{blue}{87.6}} & --& -- \\ \midrule
\cellcolor[HTML]{FFFACD}\texttt{\textbf{CLEAR}} (ours) & \cellcolor[HTML]{FFFACD}Cross-Transformers & \cellcolor[HTML]{FFFACD}\textbf{\textcolor{red}{87.2}} & \cellcolor[HTML]{FFFACD}\textbf{\textcolor{red}{91.0}}& \cellcolor[HTML]{FFFACD}\textbf{\textcolor{blue}{88.2}}& \cellcolor[HTML]{FFFACD}\textbf{\textcolor{blue}{89.8}} & \cellcolor[HTML]{FFFACD}\textbf{\textcolor{red}{81.6}}& \cellcolor[HTML]{FFFACD}\textbf{\textcolor{red}{81.2}}& \cellcolor[HTML]{FFFACD}\textbf{\textcolor{red}{83.0}} & \cellcolor[HTML]{FFFACD}\textbf{\textcolor{red}{87.9}}& \cellcolor[HTML]{FFFACD}\textbf{\textcolor{red}{85.9}}& \cellcolor[HTML]{FFFACD}\textbf{\textcolor{red}{90.0}} \\ \bottomrule
\end{tabular}}
\\
\footnote[1]{} \footnotesize Results are obtained by re-implementing the original source codes
\caption{Comparison of \texttt{\textbf{CLEAR}} with other state-of-the-art models on the PA100K, PETA, RAPv2, Market-1501 and UPAR2024-dev test for person attribute recognition task.} \vspace{-1em}
\label{tab:par}
\end{table*}

\subsection{Dataset and evaluation protocol}

To evaluate \texttt{\textbf{CLEAR}} for both tasks, we utilize five widely-used benchmarks: PA100K \cite{liu2017hydraplus}, PETA \cite{deng2014pedestrian}, RAPv2 \cite{li2018richly}, Market-1501 \cite{lin2019improving}, and UPAR2024 \cite{cormier2024upar}, whose statistics are reported in Table \ref{tab:stas}. For person attribute recognition, all five benchmarks are used. For the person retrieval task, only PA100K, Market-1501, PETA, and UPAR2024 are used. For UPAR2024, we evaluate performances in both recognition and retrieval tasks on the development test (dev-test) provided by the UPAR2024 challenge, for which annotations are available. For the retrieval task, we additionally report the results of our CLEAR on an official test set of the UPAR2024 challenge. It is worth noting that Market-1501 is the most widely used for the retrieval task, which mainly demonstrates the superior performance of \texttt{\textbf{CLEAR}} against other competitors.

Regarding evaluation metrics, we use mean accuracy (mA) and F1 score for person attribute recognition. For attribute-based person retrieval, mean average precision (mAP) and Rank-1 accuracy (R-1) are used. For ablation studies, R-5 and R-10 are additionally reported.

\subsection{Implemental Details} For both tasks, we use input images with a size of $256 \times 128$. In the cross-transformers backbone ($f^{par}$), for $\mathcal{F}_{s}$, we employ patch embedding to process the input person image into a sequence of tokens with a patch size of $4$ (kernel size = 4, stride = 4), resulting in $64 \times 32 = 2048$ tokens. Following this, we use $N_s = 4$ Swin Transformer blocks, where the embedding sizes $dim_{emb}^{s}$ and number of layers are set as $[128, 256, 512, 1024]$ and $[2, 2, 6, 2]$, respectively. The window size in the (shifted) window multi-head self-attention is set to $12$. For $\mathcal{F}_{v}$, we utilize patch embedding to process the input person image into a sequence, with a patch size of 14 (kernel size = 14, stride = 14), resulting in $18 \times 9 = 162$ tokens. Subsequently, $N_v = 12$ Vanilla Transformer blocks are employed, with the embedding size set to $dim_{emb}^{v}=1024$. Regarding $SVCF(\cdot)$ and $VSCF(\cdot)$ for cross-fusion, which are multi-head self-attention layers, we use an embedding size of $768$. For the retrieval task, we employ GPT \cite{radford2018improving} to obtain a soft embedding query for pseudo-description before it is encoded by $f^{text}_{ret}$. For two loss weights mentioned in Eq. \eqref{eq:loss_total}, we set $\beta_1 = 0.3$, and $\beta_1 = 0.7$ to encourage the contribution of soft embedding query through backward process. The implementation is done using the PyTorch library and trained using $1\times$ NVIDIA GeForce RTX 4090.

\subsection{Comparison with the State-of-the-Art}

\begin{figure}[!t]
\centerline{\includegraphics[width=0.43\textwidth]{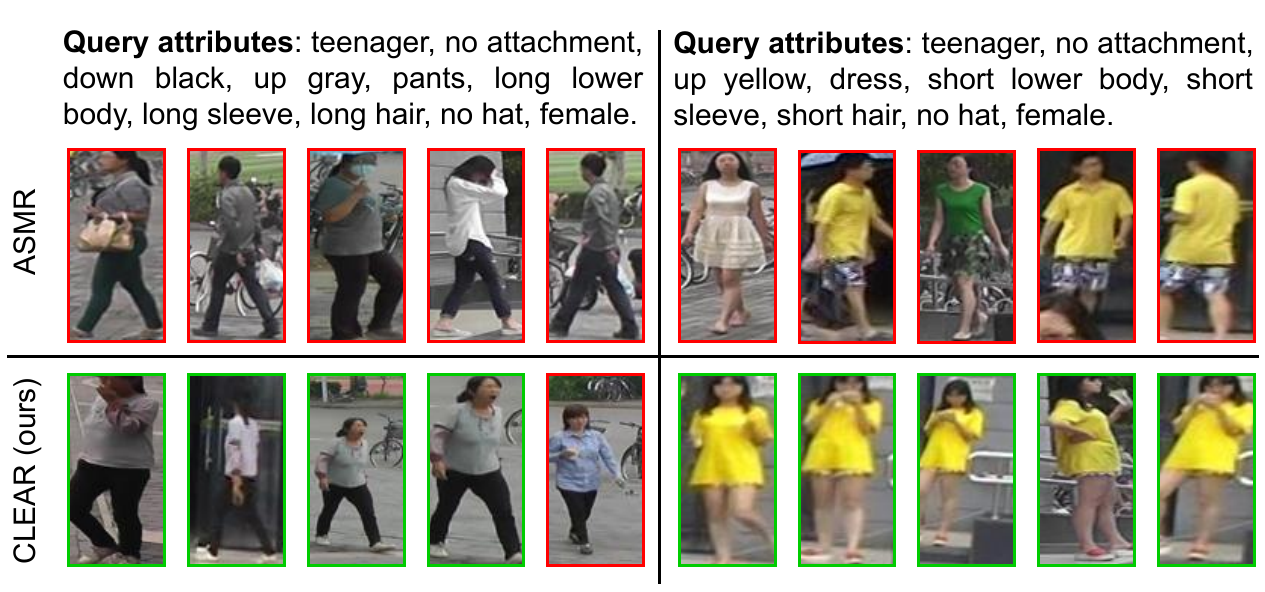}} \vspace{-0.5em}
\caption{Top five retrieval results of ASMR and \texttt{\textbf{CLEAR}} (ours).} \vspace{-0.8em}
\label{fig:retrieval}
\end{figure}

\noindent \textbf{Person attribute recognition.} In Table \ref{tab:par}, \texttt{\textbf{CLEAR}}'s performance on the PAR problem is highlighted. Our strong cross-transformer network excels at leveraging global and local-level long-range dependencies, outperforming other ResNet-based networks without human-specific modules. Notably, for PA100K and Market-1501 datasets, we surpass the UPAR model by +2.4/+0.8 and +1.5/+0.3 in mA/F1, respectively. In RAPv2, disentangled attribute feature learning helps \texttt{\textbf{CLEAR}} outperform UPAR by +0.6/+2.1 in mA/F1. For PETA, the cross-transformer network competes closely with UPAR, with only a slight difference of -0.2/-0.1 in mA/F1. On the UPAR2024 dataset, \texttt{\textbf{CLEAR}} surpasses MSCC and Strong Baseline by +1.8/+4.3 and +3.6/+3.6 in mA/F1, respectively.

\begin{table}[h]
\centering
\resizebox{0.45\textwidth}{!}{\begin{tabular}{c|cc|cc|cc}
\toprule
\multirow{2}{*}{\textbf{Methods}} & \multicolumn{2}{c}{\textbf{PETA}} & \multicolumn{2}{c}{\textbf{PA100K}} & \multicolumn{2}{c}{\textbf{Market-1501}} \\ \cmidrule{2-7}
     & R-1 & mAP & R-1 & mAP & R-1 & mAP \\ \midrule
CMCE (ICCV'17) \cite{li2017identity} & 31.7 & 26.2 & 25.8 & 13.1  
 & 35.0   & 22.8 \\
AAIPR \cite{yin2017adversarial} & 39.0 & 27.9 & -- & -- & 40.3   & 20.7 \\
AIHM (ICCV'19) \cite{dong2019person} & --   & --  & 31.3   & 17.0  & 43.3   & 24.3 \\
SAL (ECCV'20) \cite{cao2020symbiotic} & \textbf{\textcolor{blue}{39.5}}$^{*}$ & \textbf{\textcolor{blue}{36.9}}$^{*}$ & -- & -- & 49.0   & 29.8  \\
ASMR (ICCV'21) \cite{jeong2021asmr} & 24.0$^{*}$ & 26.7$^{*}$ & 31.9 & 20.6 & 49.6   & 31.0 \\
Strong baseline \cite{jia2021rethinking}   & --   & --  & 31.1 & 23.8 & 39.5  & 23.8 \\
UPAR (WACV'23) \cite{specker2023upar} & --   & --  & \textbf{\textcolor{blue}{39.5}}  & \textbf{\textcolor{blue}{30.5}} & \textbf{\textcolor{blue}{55.4}}   & \textbf{\textcolor{blue}{40.6}} \\ \midrule
\cellcolor[HTML]{FFFACD}\texttt{\textbf{CLEAR}} (ours)& \textbf{\cellcolor[HTML]{FFFACD}\textcolor{red}{48.3}} & \cellcolor[HTML]{FFFACD}\textbf{\textcolor{red}{51.8}} & \cellcolor[HTML]{FFFACD}\textbf{\textcolor{red}{46.6}} & \cellcolor[HTML]{FFFACD}\textbf{\textcolor{red}{35.9}} & \cellcolor[HTML]{FFFACD}\textbf{\textcolor{red}{56.8}} & \cellcolor[HTML]{FFFACD}\textbf{\textcolor{red}{43.1}} \\
\bottomrule
\end{tabular}}
\\
\footnote[1]{} \footnotesize Results are obtained by re-implementing the original source codes
\caption{Comparison of \texttt{\textbf{CLEAR}} with other state-of-the-art models on the PA100K, PETA, Market-1501 for attribute-based person retrieval task.} \vspace{-0.8em}
\label{tab:ret}
\end{table}

\begin{table}[]
\centering
\resizebox{0.3\textwidth}{!}{\begin{tabular}{c|cc|cc}
\toprule
\multirow{2}{*}{Method} & \multicolumn{2}{l}{\makecell{\textbf{UPAR2024} \\ \textbf{dev-test}}} & \multicolumn{2}{l}{\makecell{\textbf{UPAR2024} \\ \textbf{official-test}}}\\ \cmidrule{2-5}
 & \multicolumn{1}{l}{R-1} & \multicolumn{1}{l}{mAP} & \multicolumn{1}{l}{R-1} & \multicolumn{1}{l}{mAP} \\ \midrule
\#1 solution & - & - & 20.7 & 7.4  \\
\#2 solution & - & - & 16.1 & 6.7  \\
\#3 solution & - & - & 16.1 & 6.8  \\
ASMR & \textcolor{blue}{\textbf{27.0}}$^{*}$ & \textcolor{blue}{\textbf{19.4}}$^{*}$ & 16.8$^{*}$ & 5.3$^{*}$  \\
SAL & 23.0$^{*}$ & 16.3$^{*}$ & 13.4$^{*}$ & 3.8$^{*}$  \\
UPAR & - & - & \textcolor{blue}{\textbf{26.2}} & \textcolor{blue}{\textbf{13.4}} \\ \midrule
\cellcolor[HTML]{FFFACD}\texttt{\textbf{CLEAR}} (ours) & \cellcolor[HTML]{FFFACD}\textbf{\textcolor{red}{39.2}} & \cellcolor[HTML]{FFFACD}\textbf{\textcolor{red}{32.8}} & \cellcolor[HTML]{FFFACD}\textbf{\textcolor{red}{29.1}} & \cellcolor[HTML]{FFFACD}\textbf{\textcolor{red}{14.8}} \\
\bottomrule
\end{tabular}}
\\
\footnote[1]{} \footnotesize Results are obtained by re-implementing the original source codes
\caption{Comparison of \texttt{\textbf{CLEAR}} with other approaches on UPAR2024 dev-test and official-test within the UPAR2024 challenge for the attribute-based person retrieval task.} \vspace{-0.8em}
\label{tab:upar2024}
\end{table}

\noindent \textbf{Attribute-based person retrieval.} For attribute-based person retrieval, we conducted a comparative analysis of the \texttt{\textbf{CLEAR}} model against two adversarial learning methods, i.e. AAIPR \cite{yin2017adversarial} and SAL \cite{cao2020symbiotic}, two zero-shot learning approaches, including AIHM \cite{dong2019person} and Strong baseline \cite{jia2021rethinking}, 
 an attention learning technique CMCE \cite{li2017identity}, as well as simple yet strong frameworks such as ASMR \cite{jeong2021asmr}, and  UPAR \cite{specker2023upar}. Among the four benchmarks, Market-1501 is the most widely used, and we directly compared our results with those reported in the original studies. In the case of the PETA dataset, there were discrepancies in the statistics between our study and the SAL and ASMR studies. Consequently, we re-implemented these two methods to ensure a fair comparison. For the UPAR2024 dataset, a new dataset incorporating Market-1501, PETA, and PA100K with new annotations, we implemented two SAL and ASMR methods for comparison with \texttt{\textbf{CLEAR}}. As shown in Table \ref{tab:ret}, among the comparative methods, the UPAR model emerged as the most competitive against \texttt{\textbf{CLEAR}}. It utilizes ConvNeXt-base as a backbone with an efficient training strategy. In the case of Market-1501, we achieved a \textbf{state-of-the-art result}, surpassing the strong UPAR model by margins of 1.4 and 2.5 in terms of R-1 and mAP, respectively. For PA100K, our results significantly outperformed UPAR with improvements of 7.1 and 5.4 in terms of R-1 and mAP, respectively. Regarding PETA and UPAR2024 dev-test, since UPAR is not implemented on these two datasets, and its source code is unavailable, we compared \texttt{\textbf{CLEAR}} to the most recent SAL and ASMR models. For PETA, our performance was superior to SAL, with improvements of 8.8 and 14.9 in terms of R-1 and mAP, respectively. The results of the UPAR2024 dev-test and official test are reported in Table \ref{tab:upar2024}. For the UPAR2024 dev-test, we surpass the ASMR by a large margin of +12.2/+13.4 in R-1/mAP. In the UPAR2024 official test, we compare our performance with the top-3 competitors of the challenge and the official baseline (UPAR) \cite{cormier2024upar}. During the challenge, no teams outperform the official baseline on the large-scale, challenging UPAR2024 official test set. In contrast, $\texttt{\textbf{CLEAR}}$ surpasses UPAR by +2.9/+1.4 in R-1/mAP, achieving \textbf{state-of-the-art} on this challenging test set.

\subsection{Ablation study}

\noindent\textbf{Effect of channel-aware \& cross-fused self-attention modules.} We conducted a five-setting ablation study: 1) $\mathcal{F}^{(v)}$: Vanilla ViT; 2) $\mathcal{F}^{(s)}$: SwinT; 3) $\mathcal{F}^{(s),CASA}$: SwinT with CASA on each block; 4) $\mathcal{F}^{(v)} \oplus \mathcal{F}^{(s),CASA}$: concatenation of $\mathcal{F}^{(v)}$ and $\mathcal{F}^{(s),CASA}$; and 5) $\mathcal{F}^{(v)} \chi \mathcal{F}^{(s),CASA}$, i.e., cross-transformers backbone. Comparing $\mathcal{F}^{(v)}$ and $\mathcal{F}^{(s)}$ on PA100K and UPAR2024, Vanilla ViT excels in capturing global information for multi-class classification, while SwinT's focus on localized features may limit attention. Integrating CASA in $\mathcal{F}^{(s),CASA}$ enhances SwinT's performance, with Table \ref{tab:ablation_par} showing a +0.7 increase in mA and a slight F1 drop. Concatenating $\mathcal{F}^{(s),CASA}$ and $\mathcal{F}^{(v)}$ yields marginal improvements. Employing cross-fusion before concatenation achieves significant enhancements, with best results of 89.1\% and 87.9\% for PA100K and UPAR2024 dev-test, respectively.

\begin{table}[]
\centering
\resizebox{0.4\textwidth}{!}{\begin{tabular}{l|ccc|ccc}
\toprule
{\multirow{2}{*}{\textbf{Method}}} & \multicolumn{3}{c}{\textbf{PA100K}} & \multicolumn{3}{c}{\textbf{UPAR2024}} \\ \cmidrule{2-7}
{} & {Avg} & {mA} & {F1} & Avg & mA  & F1 \\ \midrule
\(\mathcal{F}^{(v)}\) & \textbf{\textcolor{blue}{87.9}} & 85.2 & \textbf{\textcolor{blue}{90.7}} & 87.6 & \textbf{\textcolor{blue}{85.8}} & 89.4 \\
\(\mathcal{F}^{(s)}\) & 85.7 & 82.6 & 88.8 & 86.3 & 84.3 & 88.2 \\
\(\mathcal{F}^{(s),CASA}\) & 85.9 & 83.3 & 88.6 & 86.5 & 84.9 & 88.1 \\ 
\(\mathcal{F}^{(v)}\) $\oplus$ \(\mathcal{F}^{(s),CASA}\) & 87.8 & \textbf{\textcolor{blue}{85.4}} & 90.3 & \textbf{\textcolor{blue}{87.7}} & 85.6 & \textbf{\textcolor{blue}{89.8}} \\
\midrule
\cellcolor[HTML]{FFFACD}\(\mathcal{F}^{(v)}\) $\chi$ \(\mathcal{F}^{(s),CASA}\) & \cellcolor[HTML]{FFFACD}\textbf{\textcolor{red}{89.1}} & \cellcolor[HTML]{FFFACD}\textbf{\textcolor{red}{87.2}} & \cellcolor[HTML]{FFFACD}\textbf{\textcolor{red}{91.0}} & \cellcolor[HTML]{FFFACD}\textbf{\textcolor{red}{87.9}} & \cellcolor[HTML]{FFFACD}\textbf{\textcolor{red}{85.9}} & \cellcolor[HTML]{FFFACD}\textbf{\textcolor{red}{90.0}} \\
\bottomrule
\end{tabular}}
\caption{Ablation study on PA100K test set and UPAR2024 dev-test for person attribute recognition task.} \vspace{-0.8em}
\label{tab:ablation_par}
\end{table}

\noindent\textbf{Effect of pseudo description.} To evaluate our approach, combining pseudo-descriptions from query attributes (soft embedding) and binary attributes (hard embedding), we explore three settings: 1) using only binary attribute queries (HA); 2) using word embeddings of attribute words as queries (W); and 3) relying solely on pseudo-descriptions (soft embedding). Conducted on PA100K and Market-1501, the ablation study results are summarized in Table 5. Hard embedding queries yield modest results, scoring 53.7, 70.0, and 41.1 for R-1, R-5, and mAP, respectively. Word embeddings improve performance, achieving the second-best R-5 (72.5) and mAP (43.9). Soft embedding queries yield the best mAP (44.0) and the second-best R-1 (56.6). Combining soft and hard embedding queries (HA+SP) achieves the best R-1 (56.8) and R-5 (73.3). For PA100K, HA+SP significantly outperforms HA in R-1 (46.6), R-5 (65.0), and mAP (35.9). Interestingly, HA+SP shows a notable R-1 improvement of 1.7 on PA100K but only a marginal 0.2 on Market-1501.

\begin{table}[]
\centering
\resizebox{0.35\textwidth}{!}{\begin{tabular}{c|c|ccc|ccc}
\toprule
 \multicolumn{2}{c}{\diagbox[trim=l,height=2\line]{\\ \textbf{Dataset}}{\textbf{Setting}\\ \\}} & \textbf{HA} & \textbf{W} & \textbf{SP} & \textbf{R-1} & \textbf{R-5} & \textbf{mAP}  \\ \midrule
\multirow{5}{*}{\rotatebox[origin=c]{90}{\textbf{Market-1501}}} & ResNet-50      & \ding{51} & \ding{55} & \ding{55} & 39.0& 58.5& 27.8 \\ \cmidrule{2-8}
 & \multirow{4}{*}{Cross-Trans (ours)} & \ding{51} & \ding{55} & \ding{55} & 53.7& 70.0& 41.1 \\
 & & \ding{55} & \ding{51} & \ding{55} & 55.8 & \textbf{\textcolor{blue}{72.5}} & \textbf{\textcolor{blue}{43.9}} \\
 & & \ding{55} & \ding{55} & \ding{51} & \textbf{\textcolor{blue}{56.6}} & 71.9 & \textbf{\textcolor{red}{44.0}} \\
 & & \cellcolor[HTML]{FFFACD}\ding{51} & \cellcolor[HTML]{FFFACD}\ding{55} & \cellcolor[HTML]{FFFACD}\ding{51} & \cellcolor[HTML]{FFFACD}\textbf{\textcolor{red}{56.8}} & \cellcolor[HTML]{FFFACD}\textbf{\textcolor{red}{73.3}} & \cellcolor[HTML]{FFFACD}43.1 \\ \midrule
\multirow{5}{*}{\rotatebox[origin=c]{90}{\textbf{PA100K}}}      & ResNet-50      & \ding{51} & \ding{55} & \ding{55} & 24.1  & 41.8  & 15.1   \\ \cmidrule{2-8}
 & \multirow{4}{*}{Cross-Trans (ours)} & \ding{51} & \ding{55} & \ding{55} & 43.1& \textbf{\textcolor{blue}{64.4}}& 34.3 \\
 & & \ding{55} & \ding{51} & \ding{55} & 44.3& \textbf{\textcolor{blue}{64.4}} & 35.1 \\
 & & \ding{55} & \ding{55} & \ding{51} & \textbf{\textcolor{blue}{44.9}} & 64.1 & \textbf{\textcolor{blue}{35.2}} \\
 & & \cellcolor[HTML]{FFFACD}\ding{51} & \cellcolor[HTML]{FFFACD}\ding{55} & \cellcolor[HTML]{FFFACD}\ding{51} & \multicolumn{1}{l}{\cellcolor[HTML]{FFFACD}\textbf{\textcolor{red}{46.6}}} & \multicolumn{1}{l} {\textbf{\cellcolor[HTML]{FFFACD}\textcolor{red}{65.0}}} & \textbf{\cellcolor[HTML]{FFFACD}\textcolor{red}{35.9}} \\
 \bottomrule
\end{tabular}}
\caption{Impact of three types of query forms: hard binary attribute (\textbf{HA}), soft pseudo-description (\textbf{SP}), and word embedding for attributes (\textbf{W}) on Market-1501, PA100K datasets}
\end{table}

\noindent\textbf{Effect of embedding dimension size for retrieval.} We also explore the effects of the embedding size of encoded feature vectors produced by $f^{attr}_{ret}$. In this ablation study, we consider the hard embedding query. As shown in Table 6, the results indicate that a larger embedding size leads to better performance. This can be explained by the fact that a higher embedding size provides more information for the retrieval task.

\begin{table}[]
\centering
\resizebox{0.35\textwidth}{!}{\begin{tabular}{l|lll|lll}
\toprule
\multirow{2}{*}{\diagbox[trim=l,height=2.2\line]{\\ \textbf{Size}}{\textbf{Dataset}\\ \\}} & \multicolumn{3}{c}{\textbf{Market-1501}} & \multicolumn{3}{c}{\textbf{PA100K}} \\ \cmidrule{2-7}
 & R-1  & R-5  & mAP  & R-1  & R-5  & mAP  \\ \midrule
64   & 52.1 & 69.2 & 38.2 & 42.3 & 62.5 & 33.1 \\
128  & 53.7 & 70.0 & 41.1 & 43.1 & 64.4 & 34.3 \\
\cellcolor[HTML]{FFFACD}256  & \cellcolor[HTML]{FFFACD}\textbf{\textcolor{red}{55.9}} & \cellcolor[HTML]{FFFACD}\textbf{\textcolor{red}{71.9}} & \cellcolor[HTML]{FFFACD}\textbf{\textcolor{red}{43.0}} & \cellcolor[HTML]{FFFACD}\textbf{\textcolor{red}{44.1}} & \cellcolor[HTML]{FFFACD}\textbf{\textcolor{red}{65.0}} & \cellcolor[HTML]{FFFACD}\textbf{\textcolor{red}{35.5}} \\ \bottomrule
\end{tabular}}
\caption{Impact of embedding sizes for person retrieval task on Market-1501 and PA100K datasets.} \vspace{-0.8em}
\end{table}

\subsection{Quanlitative Results} To showcase the retrieval task's success with our \texttt{\textbf{CLEAR}} model, we present t-SNE visualizations in Figure \ref{fig:tsne} for ablation settings. Each of the ten queries displays 20 person representations, highlighting that using attribute words or pseudo captions for margin learning enhances the separation of embedding vectors. In Figure \ref{fig:retrieval}, we compare the top-5 retrieval results of \texttt{\textbf{CLEAR}} and ASMR. While ASMR exhibits some confusion between genders with similar attributes, \texttt{\textbf{CLEAR}} produces more accurate results, closely matching the given queries.

\section{Conclusion}

In this study, we present \texttt{\textbf{CLEAR}}, a unified model for two human-centric tasks: person attribute recognition and attribute-based person retrieval. The proposed \texttt{\textbf{CLEAR}} model includes a robust cross-transformers backbone, exploiting global-level and local-level long-range dependencies that facilitate improved person attribute recognition. A simple yet effective strategy is introduced to adapt to the retrieval task, incorporating concepts such as a combination of a soft embedding query and a hard embedding query. Subsequently, an efficient margin learning strategy helps the unified model obtain superior results in the retrieval task. Experiments on commonly-used datasets show that our \texttt{\textbf{CLEAR}} model achieves state-of-the-art performance on both tasks, significantly advancing the benchmarks for the retrieval task.

\bibliographystyle{named}
\bibliography{ijcai24}
\end{document}